\documentclass{article}
\DeclareUnicodeCharacter{FF0C}{,}

\usepackage{PRIMEarxiv}

\usepackage[utf8]{inputenc} 
\usepackage[T1]{fontenc}    
\usepackage{hyperref}       
\usepackage{url}            
\usepackage{booktabs}       
\usepackage{amsfonts}       
\usepackage{nicefrac}       
\usepackage{microtype}      
\usepackage{lipsum}
\usepackage{fancyhdr}       
\usepackage{graphicx}       
\graphicspath{{media/}}     
\usepackage{amsmath}
\usepackage[most]{tcolorbox}
\definecolor{bg}{RGB}{240,240,240}

\title{When Large Language Models Do Not Work: Online Incivility Prediction through Graph Neural Networks
}

\author{
  Zihan Chen \\
  Stevens Institute of Technology \\
  Hoboken, New Jersey, USA \\
  \texttt{zchen61@stevens.edu} \\
   \And
  Lanyu Yu \\
  Stevens Institute of Technology \\
  Hoboken, New Jersey, USA \\
  \texttt{lanyuyudevin@gmail.com} \\
}

\begin{document}
\maketitle

\begin{abstract}
Online incivility has emerged as a widespread and persistent problem in digital communities, imposing substantial social and psychological burdens on users. Although many platforms attempt to curb incivility through moderation and automated detection, the performance of existing approaches often remains limited in both accuracy and efficiency. To address this challenge, we propose a Graph Neural Network (GNN) framework for detecting three types of uncivil behavior (i.e., toxicity, aggression, and personal attacks) within the English Wikipedia community. Our model represents each user comment as a node, with textual similarity between comments defining the edges, allowing the network to jointly learn from both linguistic content and relational structures among comments. We also introduce a dynamically adjusted attention mechanism that adaptively balances nodal and topological features during information aggregation. Empirical evaluations demonstrate that our proposed architecture outperforms 12 state-of-the-art Large Language Models (LLMs) across multiple metrics while requiring significantly lower inference cost. These findings highlight the crucial role of structural context in detecting online incivility and address the limitations of text-only LLM paradigms in behavioral prediction. All datasets and comparative outputs will be publicly available in our repository to support further research and reproducibility.\footnote{https://github.com/devinlanyu0422/WikiIncivilityGNN}
\end{abstract}


\section{Introduction}

Online communities have become integral to both organizational and public communication. However, their rapid expansion has also led to a rise in incivility and other forms of antisocial behavior. According to the Cyberbullying Research Center, nearly 33.8\% of young teenagers aged 12–17 in the United States have experienced online incivility. Abusive text and hostile online interactions, often regarded as forms of ``brutal cyber violence''~\cite{lee2018abusive, mai2019does}, pose serious threats not only to the health and reputation of online communities, but also to the mental well-being of users, contributing to anxiety, depression, and even suicidal tendencies~\cite{wright2018cyberbullying, maity2018opinion}. Consequently, curbing online incivility efficiently and effectively is an urgent priority for both online platforms and society at large. Although most online platforms have implemented mechanisms to restrain uncivil behaviors, prior studies suggest that existing systems for moderating personal attacks and cyberbullying remain weak~\cite{lee2015people}. For example, more than 30\% of user comments on major U.S. news sites still exhibit some degree of incivility~\cite{coe2014online}. Much of the research in this domain has focused on detecting uncivil messages using text classification models~\cite{zinovyeva2020antisocial, lee2018abusive}. However, these models often struggle to generalize due to evolving linguistic patterns, obfuscated spellings, context-dependence, and the absence of relational or conversational context.

To address these challenges, this paper proposes a novel Graph Neural Network (GNN) based framework for identifying uncivil comments in online communities. Specifically, we represent each user comment as a node and define edges based on textual similarity, enabling the network to jointly learn from both linguistic content and the structural relationships among user comments. During classification, this relational architecture aggregates information from both nodal features (text embeddings) and topological features (inter-comment structure) ~\cite{yao2019graph, chen2023stand}. To enhance model expressiveness, we further introduce a dynamically adjusted attention mechanism that adaptively balances these two sources of information during message passing. To evaluate our model, we use the \textit{Wikipedia Detox Project} dataset~\cite{wulczyn2017ex}, which contains an annotated subset of 100,000 Wikipedia comments labeled across three key dimensions of online incivility: (1) \emph{personal attacks}, which identify direct interpersonal hostility; (2) \emph{aggression}, capturing the intensity and tone of hostile intent; and (3) \emph{toxicity}, evaluating the extent to which a comment undermines constructive dialogue. Each comment is independently labeled by approximately ten crowdworkers, providing a robust foundation for modeling online incivility. For benchmarking, we compare our GNN-based model with twelve state-of-the-art Large Language Models (LLMs), which have demonstrated remarkable success in text-based understanding tasks. Experimental results show that our model consistently outperforms these LLMs across multiple evaluation metrics (e.g., AUC, accuracy, precision, recall, F1 score), while achieving substantially lower training and inference costs. Robustness tests further confirm that both topological and nodal features contribute significantly to our model’s predictive performance.

Our contributions are twofold. First, we propose a novel GNN-based framework for online incivility detection that integrates both linguistic and relational structures among user comments. The proposed dynamic attention mechanism enables the model to adaptively balance nodal and topological information during message aggregation, enhancing both predictive performance and interpretability. Second, we conduct a comprehensive empirical evaluation against 12 state-of-the-art LLMs, demonstrating that our approach achieves superior performance while requiring significantly lower computational cost~\cite{bai2023qwen, AnthropicC3, achiam2023gpt, qi2024benchmarking, wang2025crowdvlm}. Although LLMs have shown strong capabilities in text-based tasks, they remain computationally expensive and often fail to capture the structural dynamics that underlie social interactions. By releasing our code and model outputs, this paper offers practical value by enabling online communities to more effectively identify and mitigate uncivil behavior, thereby contributing to healthier and more constructive online environments.

The remainder of this paper is organized as follows. In the next section, we review related work on online incivility detection and graph neural network methods. We then describe our proposed graph-based framework, detailing the graph construction and dynamic attention mechanism. Following this, we present the experimental setup, data description, and results. We conclude the paper by pinpointing the potential limitations and proposing avenues for future research.

\section{Related Work}

Our study intersects two main threads of the literature: (1) online incivility and toxic behavior detection, and (2) graph neural networks implementations in online communities. In this session, we review both areas and highlight the gaps that motivate our proposed approach.

A large body of research has examined methods for detecting uncivil or antisocial behavior in online environments. Early approaches relied on lexicon-based features and traditional machine learning classifiers, using curated abusive word lists or character $n$-grams to identify harmful content~\cite{lee2018abusive}. While effective for explicit forms of abuse, these methods perform poorly when confronted with obfuscated language, contextual ambiguity, or evolving slang. To improve robustness, subsequent studies introduced large-scale annotated datasets and neural architectures. Wulczyn et al.\ released the Wikipedia Detox corpus and developed scalable models for detecting personal attacks on Wikipedia talk pages~\cite{wulczyn2017ex}, establishing one of the most influential benchmarks in toxicity and incivility research. Similarly, Sadeque et al.\ proposed neural models for fine-grained incivility (e.g., name-calling, vulgarity) in news comments, demonstrating that neural approaches outperform linear baselines for subtle forms of hostility~\cite{sadeque2019incivility}. Researchers have provided a comprehensive empirical evaluation of deep learning models for antisocial behavior detection, emphasizing both the progress achieved and the persistent challenges surrounding generalization and interpretability~\cite{zinovyeva2020antisocial, vu2025benchmarking}. More recent work has applied transformer-based architectures, including BERT variants and commercial toxicity APIs, to classify abusive or toxic content~\cite{kamphuis2024tiny}. Although these models achieve strong predictive performance, they rely solely on textual features, incur high computational cost, and do not explicitly encode relational or conversational context. As a result, text-only approaches remain vulnerable to adversarial misspellings, emerging linguistic patterns, and structural nuances in online interactions.

In parallel, graph neural networks (GNN) have emerged as powerful frameworks for learning from relational data, offering a natural way to incorporate the structural dependencies inherent in online communities. Foundational architectures such as Graph Convolutional Networks~\cite{kipf2016semi} and inductive models like GraphSAGE~\cite{hamilton2017inductive} demonstrated that aggregating information from neighboring nodes enables expressive representation learning on large, sparsely connected graphs. Building on these principles, Yao et al.\ introduced TextGCN, which constructs a global word–document graph for text classification and shows that graph-based representations can outperform traditional deep learning models~\cite{yao2019graph}. GNNs have since been applied to diverse NLP and social computing tasks, including document classification, recommendation, misinformation detection, and community modeling ~\cite{chen2023chatgpt, gao2023models}. Within the domain of toxic and hateful behavior detection, graph-based approaches have shown particular promise \cite{ sun2025objective, ShenSunQi2025}. Duong et al.\ proposed HateNet, a GCN-based architecture that constructs a semantic similarity graph over tweets, demonstrating improved hate speech detection performance relative to sequential baselines~\cite{duong2022hatenet}. Other studies integrate follower networks, user interaction graphs, or community structures to detect abusive users, highlighting the value of incorporating relational and user-level information into predictive models~\cite{samtani2020proactively, zheng2024automate, li2021neighbours, wich2021explainable}. Extensions of these approaches leverage attention mechanisms or multi-modal fusion strategies to enhance interpretability and capture dependencies within online ecosystems.

Although these studies illustrate the potential of graph neural networks for modeling antisocial behavior, several gaps remain. Most prior work focuses on user-level prediction or employs simple graph constructions, with limited attention to dynamically weighting textual versus structural signals. Furthermore, existing research does not systematically compare graph-based models with the latest large language models under consistent experimental settings. Our study addresses these gaps by proposing a comment-level GNN architecture that constructs a semantic similarity graph over Wikipedia discussion comments and introduces a dynamically adjusted attention mechanism to balance nodal (textual) and topological (structural) information during aggregation. We evaluate our model on three complementary dimensions of incivility (i.e., personal attacks, aggression, and toxicity) using high-quality, crowd-sourced annotations from the English Wikipedia community. Finally, we provide a comprehensive empirical comparison against twelve state-of-the-art LLM baselines, demonstrating that our proposed architecture not only achieves superior predictive performance across all metrics but also operates at substantially lower computational cost. These findings show that structure-aware models can outperform large-scale language models in the task of online incivility detection.

\begin{figure}[t]
  \centering
  \includegraphics[width=0.99\linewidth]{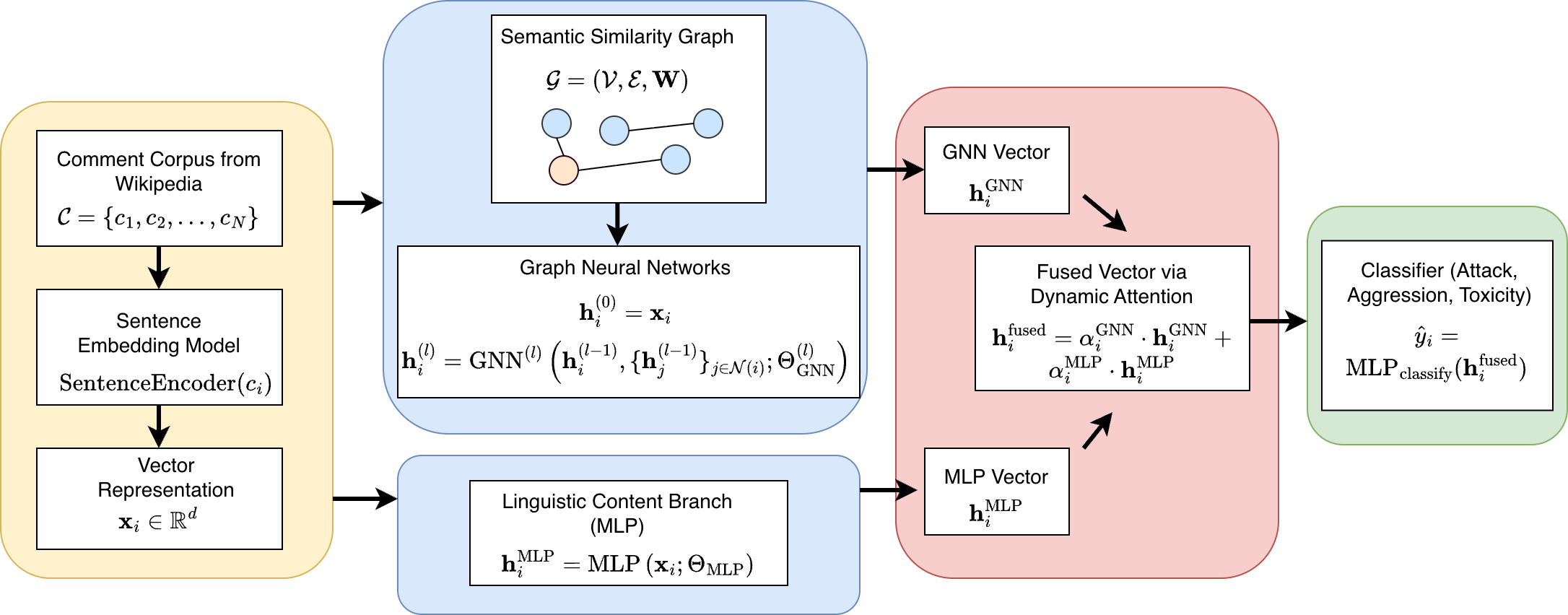}
  \caption{The high-level architecture of the proposed framework. Semantic embeddings are first generated via Sentence-BERT to construct a similarity graph. The model utilizes dual branches (GNN and MLP) fused via dynamic attention to classify comments into incivility categories.}
  \label{fig:architecture}
\end{figure}

\section{Method}

We formulate the online incivility detection as a node classification problem over three related behavioral dimensions: personal attacks, aggression, and toxicity. Given a corpus of user comments $\mathcal{C} = \{c_1, c_2, \ldots, c_N\}$, our objective is to learn three task-specific prediction functions $f_t: \mathcal{C} \rightarrow \{0, 1\}$ for $t \in \{\text{attack}, \text{aggression}, \text{toxicity}\}$, where each function determines whether a comment exhibits the corresponding form of incivility. Unlike conventional text classification approaches that treat comments as independent instances, we explicitly model the relational structure among comments through graph-based representations, enabling the joint exploitation of both linguistic content and structural context. Figure~\ref{fig:architecture} illustrates the high-level architecture of our proposed framework. In the remainder of this section, we introduce how we (1) construct the graph based on semantic similarity, (2) perform information aggregation via dynamic attention, and (3) conduct incivility classification with the corresponding loss function.

\subsection{Graph Construction via Semantic Similarity}

\textbf{Semantic Embedding Generation.} The first step of graph construction is to obtain dense vector representations of textual content. We employ a pre-trained Sentence-BERT model (all-mpnet-base-v2) \cite{reimers2019sentence}, which has been fine-tuned on over one billion sentence pairs for semantic similarity tasks. For each comment $c_i \in \mathcal{C}$, we generate a fixed-dimensional embedding vector as follows:

\begin{equation}
\mathbf{x}_i = \text{SentenceEncoder}(c_i) \in \mathbb{R}^{d}
\end{equation}

where $d = 768$ represents the dimensionality of the embedding space. The encoder applies mean-pooling over the final transformer layer outputs, producing contextually rich representations that capture semantic nuances beyond surface-level lexical features. To facilitate subsequent cosine similarity computations, we apply L2 normalization to ensure $\|\mathbf{x}_i\|_2 = 1$ for all nodes, transforming the embedding space into a unit hypersphere where inner products directly correspond to cosine similarities.

\textbf{Pairwise Similarity Computation.} Based on the embeddings, we construct task-specific comment graphs $\mathcal{G} = (\mathcal{V}, \mathcal{E}, \mathbf{W})$ for both training and test partitions, where the node set $\mathcal{V} = \{v_1, \ldots, v_N\}$ corresponds to comments, the edge set $\mathcal{E}$ encodes semantic relationships, and $\mathbf{W}: \mathcal{E} \rightarrow \mathbb{R}^+$ assigns similarity-based weights. Specifically, for each pair of normalized comment embeddings $(\mathbf{x}_i, \mathbf{x}_j)$, we compute the cosine similarity as:

\begin{equation}
s_{ij} = \mathbf{x}_i^\top \mathbf{x}_j = \frac{\mathbf{x}_i^\top \mathbf{x}_j}{\|\mathbf{x}_i\|_2 \|\mathbf{x}_j\|_2}
\end{equation}

where the equality holds due to L2 normalization. The resulting similarity matrix $\mathbf{S} \in \mathbb{R}^{N \times N}$ quantifies semantic relatedness across all comment pairs, where high similarity values indicate that two comments share similar linguistic patterns, topical content, or rhetorical structures. We set diagonal elements $s_{ii} = 1$ to include self-loops, ensuring that message-passing operations aggregate information from both node itself and its neighboring nodes.

\textbf{Edge Formation with Connectivity Constraint.} An edge $(v_i, v_j) \in \mathcal{E}$ is established if the pairwise similarity exceeds a predefined threshold $\tau$, with the corresponding edge weight defined as $w_{ij} = s_{ij}$.

\begin{equation}
\mathcal{E} = \{(v_i, v_j) : s_{ij} > \tau, \, i \neq j\}
\end{equation}

In our experiments, we set $\tau = 0.9$ to ensure that edges represent strong semantic relationships, filtering out spurious connections that could introduce noise during message passing. To prevent isolated nodes or weakly connected components that could impede information propagation, we enforce a minimum connectivity constraint: for each node $v_i$, if fewer than $k_{\min}$ neighbors satisfy the threshold condition, we connect $v_i$ to its top-$k_{\min}$ nearest neighbors according to similarity scores:

\begin{equation}
\mathcal{N}(v_i) = \begin{cases}
\{v_j : s_{ij} > \tau\}, & \text{if } |\{v_j : s_{ij} > \tau\}| \geq k_{\min} \\
\text{top-}k_{\min}(\{v_j\}_{j \neq i}, s_{ij}), & \text{otherwise}
\end{cases}
\end{equation}

where $\mathcal{N}(v_i)$ denotes the neighbor set of node $v_i$. This constraint ensures that all nodes participate in message passing, enabling even comments with atypical linguistic characteristics to receive contextual information from structurally proximate nodes. The resulting graph is made undirected by adding reverse edges $(v_j, v_i)$ for each $(v_i, v_j) \in \mathcal{E}$, enforcing symmetric information flow during aggregation.

\subsection{Hybrid Graph Neural Network Architecture}

We propose a hybrid neural architecture that integrates two complementary information pathways: a graph neural network branch that exploits topological structure and inter-comment relationships, and a multi-layer perceptron branch that independently processes nodal features. These branches are combined through a learnable attention mechanism that dynamically balances their contributions.

\subsubsection{Graph Neural Network Branch}

The GNN branch aggregates information from semantically similar comments through multi-layer message passing. Given the initial node features $\mathbf{h}_i^{(0)} = \mathbf{x}_i$, where $\mathbf{x}_i$ is the comment embedding from Sentence-BERT, we apply $L$ layers of graph neural network operations to produce the final GNN representation:

\begin{equation}
\mathbf{h}_i^{(l)} = \text{GNN}^{(l)}\left(\mathbf{h}_i^{(l-1)}, \{\mathbf{h}_j^{(l-1)}\}_{j \in \mathcal{N}(i)}; \Theta_{\text{GNN}}^{(l)}\right)
\end{equation}

for $l = 1, \ldots, L$, where $\Theta_{\text{GNN}}^{(l)}$ denotes the learnable parameters at layer $l$, and $\mathcal{N}(i)$ represents the neighbor set of node $v_i$. The GNN operation aggregates features from neighboring nodes weighted by their semantic similarity, enabling the model to learn contextual representations that incorporate information from structurally related comments. The choice of GNN architecture can be any differentiable message-passing model (e.g., GAT, GCN) that allows the model to capture diverse relational patterns simultaneously. Residual connections are incorporated between consecutive layers to enhance gradient propagation in deep architectures. The final output $\mathbf{h}_i^{\text{GNN}} = \mathbf{h}_i^{(L)}$ represents the graph-aware representation of node $v_i$.

\subsubsection{Multi-Layer Perceptron Branch}

The MLP branch processes node features independently without leveraging graph structure, serving as a complementary pathway that captures node-specific linguistic characteristics:

\begin{equation}
\mathbf{h}_i^{\text{MLP}} = \text{MLP}\left(\mathbf{x}_i; \Theta_{\text{MLP}}\right)
\end{equation}

where $\Theta_{\text{MLP}}$ denotes the learnable parameters of the multi-layer perceptron. The MLP consists of fully connected layers with non-linear activations, transforming the input embedding $\mathbf{x}_i$ to a hidden representation $\mathbf{h}_i^{\text{MLP}}$. This branch enables the model to identify incivility based solely on textual content, complementing the structural reasoning provided by the GNN branch. By maintaining dimensional consistency with the GNN branch output, we facilitate balanced fusion in subsequent stages.

\subsubsection{Dynamic Attention-Based Fusion}

To integrate information from both branches, we introduce a dynamic attention mechanism that learns instance-specific importance weights. We first concatenate the representations from both pathways:

\begin{equation}
\mathbf{h}_i^{\text{concat}} = \text{CONCAT}\left(\mathbf{h}_i^{\text{GNN}}, \mathbf{h}_i^{\text{MLP}}\right)
\end{equation}

The attention mechanism then computes normalized importance scores for each branch through a two-layer neural network:

\begin{equation}
\left[\alpha_i^{\text{GNN}}, \alpha_i^{\text{MLP}}\right] = \text{Softmax}\left(\text{ATT}\left(\mathbf{h}_i^{\text{concat}}; \Theta_{\text{ATT}}\right)\right)
\end{equation}

where $\Theta_{\text{ATT}}$ represents the attention network parameters, and $\alpha_i^{\text{GNN}}, \alpha_i^{\text{MLP}} \in [0,1]$ satisfy $\alpha_i^{\text{GNN}} + \alpha_i^{\text{MLP}} = 1$. These learned weights represent the relative importance of structural versus content-based features for classifying each comment. The attention mechanism enables the model to emphasize graph structure for comments whose incivility is contextually dependent (e.g., sarcasm, coded language) while prioritizing textual content for comments with explicit offensive markers. We then compute the attention-weighted fusion:

\begin{equation}
\mathbf{h}_i^{\text{fused}} = \alpha_i^{\text{GNN}} \cdot \mathbf{h}_i^{\text{GNN}} + \alpha_i^{\text{MLP}} \cdot \mathbf{h}_i^{\text{MLP}}
\end{equation}

This weighted combination allows the model to adaptively balance the two information sources on a per-instance basis, enhancing both predictive performance and interpretability through analysis of the learned attention patterns.

\subsubsection{Classification and Loss Function}

The fused representation is passed through a multi-layer classification network to produce the final prediction:

\begin{equation}
\hat{y}_i = \sigma\left(\text{MLP}_{\text{classify}}\left(\mathbf{h}_i^{\text{fused}}; \Theta_{\text{classify}}\right)\right)
\end{equation}

where $\sigma(\cdot)$ denotes the sigmoid activation function, $\Theta_{\text{classify}}$ represents the classifier parameters, and $\hat{y}_i \in (0,1)$ is the predicted probability that comment $i$ exhibits the target form of incivility. The model predictions $\hat{y}_i$ are then compared with ground-truth labels $y_i$ for personal attacks, aggression, and toxicity detection independently, using binary cross-entropy loss. The loss is then backpropagated to train the model end-to-end using the Adam optimizer. We evaluate model performance using five complementary metrics including AUC-ROC, F1-Score, Precision, Recall, and Accuracy. Additionally, we monitor averaged attention weights $\{\alpha^{\text{GNN}}, \alpha^{\text{MLP}}\}$ to interpret branch contributions and compute train-test AUC gaps to quantify overfitting. In the next section, we introduce our experimental settings and analyze model performance across these metrics.

\section{Experiment}

To evaluate the performance of our proposed model, we implement it on the Wikipedia Detox dataset. This section presents the dataset, experimental setup, and empirical results. The results demonstrate that our model consistently outperforms all benchmark methods across multiple evaluation metrics.

\subsection{Data}

We employ the \textit{Wikipedia Detox Project} dataset, a large-scale corpus of discussion comments from the English Wikipedia community, annotated across three dimensions of online incivility: \textit{toxicity}, \textit{aggression}, and \textit{personal attacks}. The dataset comprises 196{,}384 unique comments, each independently evaluated by approximately 10–12 crowdworkers (mean = 10.9, SD = 0.9), making it one of the most extensively labeled resources for studying uncivil online discourse. Table~\ref{tab:dataset_stats} summarizes the key statistics for each category.

\begin{table}[!ht]
\centering
\caption{Dataset statistics across three incivility dimensions.}
\label{tab:dataset_stats}
\begin{tabular}{lrrrrr}
\toprule
\textbf{Category} & \textbf{Comments} & \textbf{Annotations} & \textbf{Positive (\%)} & \textbf{Agreement} & \textbf{Avg. Length} \\
\midrule
Toxicity & 159{,}463 & 1{,}598{,}289 & 11.5 & 0.918 & 447 \\
Aggression & 115{,}705 & 1{,}365{,}217 & 14.7 & 0.890 & 461 \\
Personal Attack & 115{,}705 & 1{,}365{,}217 & 13.5 & 0.901 & 461 \\
\bottomrule
\end{tabular}
\end{table}

Inter-annotator reliability is high across all categories, with mean agreement scores of 0.918 (toxicity), 0.890 (aggression), and 0.901 (personal attacks). Over half of all comments (48–59\%) achieved perfect annotator consensus, and 79–88\% showed high agreement ($\geq$0.8), suggesting that while incivility can be subjective, it is reliably identifiable by human annotators. We apply majority voting to derive ground-truth labels, resolving ties in favor of the positive class to ensure conservative detection.

The dataset exhibits a natural class imbalance, with positive labels comprising 11.5\% (toxicity), 14.7\% (aggression), and 13.5\% (personal attacks) of the total instances. This imbalance reflects the real-world prevalence of uncivil discourse in Wikipedia discussions and presents a realistic evaluation setting. The imbalance ratios range from 5.8:1 to 7.7:1. To address this issue, we construct a balanced version of the dataset containing roughly 50\% uncivil and 50\% civil comments. The data are then split into training, validation, and test sets with an 8:1:1 ratio. Furthermore, Comments originate from two namespace types: user talk pages (55–59\%) and article discussion pages (41–45\%). Notably, toxic comments are significantly shorter than civil ones across all categories, with an average length of 389 characters compared to 464 characters for civil comments, suggesting that hostile remarks are generally more concise and direct.

\subsection{Model Configuration and Evaluation Metrics}

We implement our hybrid GNN-MLP architecture using PyTorch 2.0+ and PyTorch Geometric for graph operations. The model is trained for each task (personal attacks, aggression, and toxicity) to allow task-specific graph structures and parameter optimization. For the GNN branch, we employ Graph Attention Networks (GAT) with three layers ($L=3$), using multi-head attention with $K=3$ heads for intermediate layers and single-head attention for the final layer. Both GNN and MLP branches produce representations with hidden dimension $d_h = 256$. The MLP branch consists of two fully connected layers, and the classification network contains three layers with progressive dimensionality reduction (256 $\rightarrow$ 128 $\rightarrow$ 1). We train models using the Adam optimizer and dropout in MLP components to prevent overfitting, while batch normalization is applied after each graph convolution layer. Models are trained for a maximum of 500 epochs with early stopping based on validation set performance. All experiments are conducted on NVIDIA GPUs with full-graph batch training to leverage the complete graph structure during each optimization step.

To comprehensively assess model performance on imbalanced binary classification tasks, we employ five complementary metrics. All metrics are computed using scikit-learn with default parameters, and models are evaluated on held-out test sets that remain fixed across all comparative experiments to ensure fair comparison with baseline methods.

\begin{itemize}
\item \textbf{AUC-ROC}: Measures the model's ability to discriminate between positive and negative classes across all decision thresholds, providing a threshold-independent evaluation of ranking quality. This serves as our primary evaluation metric due to its robustness to class imbalance.

\item \textbf{F1-Score}: The harmonic mean of precision and recall, computed as $F_1 = 2 \cdot \frac{\text{Precision} \cdot \text{Recall}}{\text{Precision} + \text{Recall}}$, providing a balanced measure that accounts for both false positives and false negatives.

\item \textbf{Precision}: The fraction of positive predictions that are correct, $\text{Precision} = \frac{TP}{TP + FP}$, where $TP$ denotes true positives and $FP$ denotes false positives. This metric indicates the model's ability to avoid false alarms in content moderation scenarios.

\item \textbf{Recall}: The fraction of actual positive instances correctly identified, $\text{Recall} = \frac{TP}{TP + FN}$, where $FN$ denotes false negatives. This metric measures the model's sensitivity to incivil content and its ability to minimize missed detections.

\item \textbf{Accuracy}: Overall classification correctness, $\text{Accuracy} = \frac{TP + TN}{TP + TN + FP + FN}$, where $TN$ denotes true negatives. While reported for completeness, this metric may be less informative under severe class imbalance.
\end{itemize}

\subsection{Results}

In this section, we present the empirical results of our proposed GNN-based framework compared to twelve state-of-the-art Large Language Models (LLMs). We evaluate the models across three distinct dimensions of online incivility: personal attacks, aggression, and toxicity. Across all tasks and metrics (AUC, accuracy, precision, recall, and F1), our model achieves consistently strong performance and, in particular, attains the highest AUC on every task (Tables~\ref{tab:attack}–\ref{tab:toxicity}). These gains indicate that incorporating relational structure among comments substantially improves the model’s ability to distinguish civil from uncivil content, especially for borderline cases.

\begin{table}[h]
\centering
  \caption{Model Performance on Attack Detection}
  \label{tab:attack}
  \begin{tabular}{lccccc}
    \toprule
    Model & Acc & Pre & Recall & F1 & AUC\\
    \midrule 
    nova-lite & 0.807 & 0.927 & 0.608 & 0.734 & 0.800 \\
    llama3.2-11b & 0.867 & 0.782 & 0.972 & 0.867 & 0.899 \\
    nova-pro & 0.884 & 0.865 & 0.871 & 0.868 & 0.921 \\
    qwen3-32b & 0.890 & 0.881 & 0.871 & 0.876 & 0.929 \\
    claude-sonnet-3.7 & 0.880 & 0.861 & 0.872 & 0.866 & 0.930 \\
    llama4-scout-17b & 0.895 & 0.879 & 0.886 & 0.883 & 0.931 \\
    llama3.3-70b & 0.872 & 0.803 & 0.945 & 0.868 & 0.935 \\
    llama4-maverick-17b & 0.863 & 0.788 & 0.946 & 0.860 & 0.938 \\
    mistral-large & 0.878 & 0.804 & 0.961 & 0.876 & 0.939 \\
    claude-sonnet-4 & 0.883 & 0.891 & 0.839 & 0.865 & 0.941 \\
    qwen3-235b & 0.885 & 0.824 & 0.945 & 0.880 & 0.942 \\
    nova-premier & 0.883 & 0.821 & 0.905 & 0.861 & 0.944 \\
    \textbf{Our Model} & 0.887 & 0.862 & 0.889 & 0.875 & \textbf{0.957} \\
  \bottomrule
\end{tabular}
\end{table}

\paragraph{Personal Attack Detection:}
Table~\ref{tab:attack} reports the performance of all models on detecting personal attacks. Our model achieves the highest AUC (0.957), improving over the strongest LLM baseline (nova-premier, AUC = 0.944) by 1.3 percentage points. Our results also shows a general pattern that more recent and larger LLM variants achieve stronger performance (e.g., nova-premier outperforming nova-pro and nova-lite), indicating that scale and recency often translate into improved predictive ability. Notably, some LLMs (e.g., nova-lite) exhibit very high precision (0.927) at the cost of substantially reduced recall (0.608), suggesting a conservative classification strategy that misses many attacks. By contrast, our model maintains a more balanced precision–recall profile (precision = 0.862, recall = 0.889), reflecting its ability to detect a larger fraction of personal attacks without incurring excessive false positives.

\begin{table}[h]
\centering
  \caption{Model Performance on Aggression Detection}
  \label{tab:aggression}
  \begin{tabular}{lccccc}
    \toprule
    Model & Acc & Pre & Recall & F1 & AUC\\
    \midrule
    nova-lite & 0.782 & 0.910 & 0.564 & 0.696 & 0.861 \\
    llama3.2-11b & 0.861 & 0.796 & 0.926 & 0.856 & 0.910 \\
    llama4-maverick-17b & 0.868 & 0.825 & 0.896 & 0.859 & 0.925 \\
    llama3.3-70b & 0.880 & 0.831 & 0.919 & 0.873 & 0.931 \\
    llama4-scout-17b & 0.876 & 0.835 & 0.901 & 0.867 & 0.933 \\
    qwen3-235b & 0.868 & 0.842 & 0.868 & 0.855 & 0.935 \\
    qwen3-32b & 0.864 & 0.902 & 0.781 & 0.837 & 0.938 \\
    nova-pro & 0.871 & 0.864 & 0.840 & 0.852 & 0.940 \\
    mistral-large & 0.880 & 0.832 & 0.917 & 0.873 & 0.945 \\
    nova-premier & 0.874 & 0.890 & 0.799 & 0.842 & 0.948 \\
    claude-sonnet-4 & 0.873 & 0.899 & 0.806 & 0.850 & 0.950 \\
    claude-sonnet-3.7 & 0.888 & 0.867 & 0.887 & 0.877 & 0.953 \\
    \textbf{Our Model} & 0.904 & 0.896 & 0.889 & 0.892 & \textbf{0.962} \\
  \bottomrule
\end{tabular}
\end{table}

\paragraph{Aggression Detection:}
For aggression detection (Table~\ref{tab:aggression}), our model provides a clear improvement over all baselines. It achieves the highest AUC (0.962) and accuracy (0.904), exceeding the best LLM (claude-sonnet-3.7, AUC = 0.953, accuracy = 0.888). Our model also yields the highest F1 score (0.892), outperforming the strongest LLM baseline (0.877 for claude-sonnet-3.7) by 1.5 percentage points. While some models, such as nova-lite and nova-premier, achieve very high precision (0.910 and 0.890, respectively), they do so with markedly lower recall (0.564 and 0.799), again indicating a tendency to under-flag aggressive content. In contrast, our model simultaneously preserves high precision (0.896) and recall (0.889), suggesting that leveraging the graph structure helps capture more subtle and context-dependent forms of aggression that purely text-based models overlook.

\begin{table}[h]
\centering
  \caption{Model Performance on Toxicity Detection}
  \label{tab:toxicity}
  \begin{tabular}{lccccc}
    \toprule
    Model & Acc & Pre & Recall & F1 & AUC\\
    \midrule
    nova-lite & 0.832 & 0.921 & 0.713 & 0.804 & 0.841 \\
    llama3.2-11b & 0.875 & 0.809 & 0.972 & 0.883 & 0.917 \\
    llama4-scout-17b & 0.893 & 0.872 & 0.916 & 0.894 & 0.937 \\
    qwen3-235b & 0.883 & 0.835 & 0.949 & 0.888 & 0.944 \\
    llama3.3-70b & 0.895 & 0.857 & 0.943 & 0.898 & 0.947 \\
    nova-pro & 0.896 & 0.887 & 0.900 & 0.893 & 0.954 \\
    llama4-maverick-17b & 0.893 & 0.866 & 0.925 & 0.894 & 0.954 \\
    mistral-large & 0.881 & 0.823 & 0.965 & 0.888 & 0.954 \\
    nova-premier & 0.889 & 0.839 & 0.943 & 0.888 & 0.956 \\
    qwen3-32b & 0.895 & 0.900 & 0.884 & 0.892 & 0.958 \\
    claude-sonnet-3.7 & 0.893 & 0.850 & 0.950 & 0.897 & 0.960 \\
    claude-sonnet-4 & 0.899 & 0.906 & 0.886 & 0.896 & 0.963 \\
    \textbf{Our Model} & 0.913 & 0.919 & 0.901 & 0.910 & \textbf{0.970} \\
  \bottomrule
\end{tabular}
\end{table}

\paragraph{Toxicity Detection:}
Toxicity detection exhibits a similar pattern (Table~\ref{tab:toxicity}). Our model attains the highest AUC (0.970) and accuracy (0.913), improving upon the strongest LLM baseline (claude-sonnet-4, AUC = 0.963, accuracy = 0.899). It also reaches the best F1 score (0.910), outperforming the best LLM (0.898 for llama3.3-70b) by 1.2 percentage points. Although some LLMs achieve slightly higher recall (e.g., llama3.2-11b with recall = 0.972), they typically trade off precision, whereas our model retains both high precision (0.919) and recall (0.901). This balanced performance is particularly important for practical deployment, where both missed toxic comments and over-flagging benign content are costly.

\paragraph{Summary:}
Comparing across tasks, personal attack detection appears to be the most challenging, with lower F1 scores overall relative to aggression and toxicity. Nonetheless, our model consistently achieves the highest AUC on all three tasks and the highest F1 score on two of them (aggression and toxicity), while remaining competitive on personal attacks. The pattern of results suggests that modeling relational structure among comments is especially beneficial for dimensions such as aggression and toxicity, which rely heavily on conversational context and tone. Taken together, these findings support our central claim that integrating nodal text features with graph-based topological information yields more robust and reliable detection of online incivility than text-only LLM approaches.

\section{Discussion}

This paper proposes a graph-based framework for detecting online incivility that integrates both linguistic content and relational structure among user comments. By representing each comment as a node and connecting comments via text-based similarity edges, our model leverages Graph Neural Networks with a dynamically adjusted attention mechanism to balance nodal (textual) and topological (structural) information during message passing. This design allows the model to move beyond isolated comment classification and instead reason over the broader interaction context in which uncivil behavior emerges. Empirically, our framework achieves consistently strong performance across three core dimensions of online incivility (i.e., personal attacks, aggression, and toxicity) on the Wikipedia Detox dataset. Compared against twelve state-of-the-art LLMs, our model attains the highest AUC on all three tasks and maintain a more balanced precision–recall trade-off. These results indicate that incorporating structural context meaningfully improves the detection of uncivil comments, particularly in borderline or ambiguous cases where textual cues alone may be insufficient. At the same time, the model is substantially more computationally efficient than large LLMs, suggesting that graph-based approaches offer a practical and scalable alternative for content moderation in real-world platforms.

Despite these promising findings, our work has several limitations that open up opportunities for future research. First, the current graph construction relies solely on textual similarity to define edges between comments. While this choice provides a simple and interpretable relational structure, it captures only one facet of how comments relate to one another. In practice, online interactions are shaped by additional signals, such as shared authorship, reply or thread structure, temporal proximity, and users' historical behavior. Future work could incorporate these signals by creating edges between comments from the same user, encoding conversation trees, or introducing user-level nodes with attributes derived from prior behavior. Such extensions would enrich both nodal and topological features, potentially enabling the model to better capture recurring patterns of incivility and the social dynamics underlying them.

Second, our current architecture deliberately uses relatively simple neural network blocks (i.e., GAT layers and multilayer perceptrons) to demonstrate that meaningful gains can be achieved even without highly complex models. However, more expressive architectures may further improve performance. For example, transformer-based encoders could be employed to generate richer comment embeddings before graph propagation, or transformer-style graph networks could jointly model long-range textual dependencies and higher-order structural patterns. In parallel, scaling up the training data by incorporating additional platforms, larger comment corpora, or multilingual datasets could enhance robustness and generalizability across domains. Exploring the explainability of deep learning models, for example by identifying the most impactful neighbors and understanding how they influence the information propagation path, also represents a promising direction~\cite{dwivedi2023explainable, cao2023modeling}.

Overall, our findings highlight the importance of moving beyond purely text-based models and explicitly modeling the relational context in which online incivility occurs. By combining nodal text features with graph-based topological information, our framework provides a robust and efficient approach to detecting uncivil behavior. Future work that broadens the sources of structural information, adopts more advanced architectures, and scales to richer datasets has the potential to further strengthen automated moderation tools and support healthier online communities.

\bibliographystyle{unsrt}  
\bibliography{references}

\end{document}